%% file: main.tex
\documentclass{article}
\usepackage{arxiv}

\usepackage[utf8]{inputenc} 
\usepackage[T1]{fontenc}    
\usepackage{hyperref}       
\usepackage{url}            
\usepackage{booktabs}       
\usepackage{amsfonts}       
\usepackage{nicefrac}       
\usepackage{microtype}      
\usepackage{lipsum}
\usepackage{graphicx}
\graphicspath{ {./images/} }

\usepackage{graphicx}
\usepackage{hyperref}
\usepackage{amsmath}
\usepackage{algpseudocode}
\usepackage{algorithm}
\usepackage{amsfonts}
\usepackage{longtable}
\usepackage{booktabs}
\usepackage{tabularx}

\title{Prioritizing Search Space Regions in the Low Autocorrelation Binary Sequences Problem}

\newcommand\orcid[1]{\href{https://orcid.org/#1}{\includegraphics[scale=0.09]{ORCIDiD_icon128x128.png}}}

\author{ \href{https://orcid.org/0009-0000-1598-8056}{\includegraphics[scale=0.06]{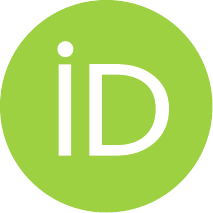}\hspace{1mm}Blaž Pšeničnik}\\
	Computer Architecture and Languages Laboratory\\
	Faculty of Electrical Engineering and Computer Science\\
	University of Maribor\\
	\texttt{blaz.psenicnik1@um.si} \\
	\And
	\href{https://orcid.org/0000-0002-7595-2845}{\includegraphics[scale=0.06]{orcid.pdf}\hspace{1mm}Borko Bošković} \\
	Computer Architecture and Languages Laboratory\\
	Faculty of Electrical Engineering and Computer Science\\
	University of Maribor\\
	\texttt{borko.boskovic@um.si} \\
    \And
	\href{https://orcid.org/0000-0002-7156-7050}{\includegraphics[scale=0.06]{orcid.pdf}\hspace{1mm}Jan Popič} \\
	Computer Architecture and Languages Laboratory\\
	Faculty of Electrical Engineering and Computer Science\\
	University of Maribor\\
	\texttt{jan.popic1@um.si} \\
 	\And
	\href{https://orcid.org/0000-0001-5864-3533}{\includegraphics[scale=0.06]{orcid.pdf}\hspace{1mm}Janez Brest} \\
	Computer Architecture and Languages Laboratory\\
	Faculty of Electrical Engineering and Computer Science\\
	University of Maribor\\
	\texttt{janez.brest@um.si} \\
}

\begin{document}
\maketitle
\begin{abstract}
Low autocorrelation binary sequences problem (LABS) is a hard combinatorial optimization challenge with important applications in communications, signal processing, and satellite navigation. This paper proposes a hybrid search framework that combines Thompson sampling with parallel self-avoiding walks to adaptively allocate computational effort across restriction classes of the LABS search space. By modeling partitions as arms in a multi-armed bandit setting, the proposed method dynamically shifts search resources toward partitions that empirically produce higher merit factors while maintaining exploration of less-sampled regions. The approach is further accelerated through GPU-parallel execution, shared posterior updates, efficient neighborhood evaluation, and a Bloom filter for cycle prevention. In addition, we use a two-stage optimization strategy that first searches constrained partitioned skew-symmetric spaces and then refines the best candidates in the unrestricted space. Experiments on long binary sequences show that the proposed method improves the previously best-known results for 35 sequence lengths in the range $450 \le L \le 527$ and for $L=573$. In particular, we report a new longest sequence with merit factor exceeding $8.0$, obtained for $L=451$. The results also show that Thompson sampling effectively prioritizes partitions with better observed performance, confirming the value of online, data-driven resource allocation in LABS optimization. Overall, the proposed framework provides a scalable and effective strategy for high-performance merit factor maximization.
\end{abstract}


\section*{Introduction}
The study of low autocorrelation binary sequences (LABS) is recognized as a highly challenging computational problem, belonging to the class of hard binary combinatorial problems. It was formally introduced in 1972 by Golay~\cite{golay72}. Earlier groundwork was laid by Littlewood~\cite{littlewood66}, a mathematician, who examined polynomials with coefficients restricted to $\pm 1$ on the unit circle in the complex plane, a problem closely related to LABS. Sequences with low autocorrelation are valuable across a range of practical contexts. In digital communications, they help distinguish signals from background noise more effectively~\cite{jedwab04,katz24} and are critical for packet detection and bit alignment~\cite{skula26}, especially for low‑power Internet of Things receivers. Beyond that, their usefulness extends to areas such as physics~\cite{bernasconi87}, chemistry, and cryptography.  A broader overview of additional applications and theoretical developments can be found in the survey literature on this topic~\cite{jedwab04}. A particularly striking application involved their role in highly accurate interplanetary radar experiments designed to test the curvature of space-time~\cite{shapiro68}. In global navigation satellite systems (GNSS), sequences with low autocorrelation play a crucial role; when combined with additional desirable properties, they are known as spreading codes~\cite{yang24a}. For example, the \textit{Global Positioning System L1 C/A signal} uses a set of $63$ distinct spreading codes, each of length $1023$~\cite{yang24a}. These types of codes are also important in low Earth orbit (LEO) satellite applications~\cite{yang24b}. In recent years, interest in this problem has grown further with the emergence of quantum computing, as such combinatorial optimization challenges are seen as promising candidates for quantum algorithms~\cite{sciorilli25,shaydulin24}.

In the literature, binary sequences are commonly analyzed using two forms of autocorrelation: periodic and aperiodic autocorrelation functions~\cite{schmidt16}. The aperiodic autocorrelation is often regarded as a more realistic characterization of sequence behavior in practical systems~\cite{kettunen97}. A binary sequence of length $L$ is defined as $S(L)=\{s_1, s_2, \ldots, s_L\}$, where element satisfies $s_i \in \{+1, -1\}$. The aperiodic autocorrelation function is given by:
\begin{equation}\label{eq:correlation}
    C_k(S)=\sum_{i=1}^{L-k} s_i s_{i+k}, \quad k \in \{1, \dots, L\}.
\end{equation}

The sequence energy is defined as $E(S)=\sum_{k=1}^{L-1} C_k^2(S)$, i.e., the sum of squared aperiodic autocorrelation values over all non-zero shifts $k$. Sequence search and design methods in the literature typically follow two principal optimization criteria. The first aims to minimize the peak sidelobe level (PSL), while the second seeks to maximize the merit factor ($F$)~\cite{mullen13}, defined as:
\begin{equation}\label{eq:merit_factor}
F(S)=\frac{L^2}{2E(S)}.
\end{equation}
These two objectives constitute a trade-off, making their simultaneous optimization generally infeasible. Accordingly, practical sequence design does not attempt to balance PSL and MF; instead, it focuses on optimizing either PSL or $F$, depending on the application and design requirements~\cite{brest25}. The goal of the LABS problem is to identify a binary sequence $S^*$ that, for a given length $L$, achieves the maximum possible merit factor $F$:
\begin{equation}\label{eq:problem}
S^* = \underset{S \in \{-1,1\}^L}{\operatorname{arg\max}} F(S).
\end{equation}
In other words, the objective is to find a sequence that maximizes the merit factor defined in Equation~\eqref{eq:merit_factor}, thereby minimizing the associated autocorrelation energy.

Since LABS is a binary combinatorial optimization problem, the size of the search space grows exponentially with the sequence length $L$, specifically as $2^L$. The landscape is characterized by an exponentially increasing number of local minima as $L$ increases, while the global minima are extremely rare, highly isolated, and sharply defined, resembling the shape of a golf hole~\cite{oliveira99}. Figure~\ref{fig:search_space} illustrates the distribution of global and local optima using a two-dimensional projection of the search space. This projection is obtained via UMAP (Uniform Manifold Approximation and Projection)~\cite{healy24}, a dimensionality reduction technique, for sequence lengths $L=12,\,15,\,17$. The values $C_k(S)$ in Equation~\ref{eq:correlation} remain unchanged if the sign of each sequence element is flipped (i.e., multiplied by $-1$) or if the sequence is reversed. Under these transformations, the sequence energy remains invariant~\cite{packebusch16}.  If alternating elements of the sequence are complemented, correlations with odd indices $k$ remain unchanged, while correlations with even indices only change sign. Hence, all sequences of length $L$ can be can be grouped into eight mutually equivalent classes. Consequently, the number of nonequivalent sequences is slightly greater than $2^{(L-3)}$~\cite{packebusch16}. In Figure~\ref{fig:search_space}, these symmetry classes are visible as distinct clusters for each sequence length.

\input{FIG_SEARCH_SPACE}

Due to the exponential growth of the problem’s search space, its reduction can be highly advantageous. A binary sequence of odd length $L=2k+1$ is called skew-symmetric~\cite{golay72} if it satisfies the following condition:
\begin{equation}\label{eq:skew_symmetry}
s_{(k+1)+i}=(-1)^i s_{(k+1)-i},\quad i = 1,2,\,\cdots,k.
\end{equation}
This constraint effectively reduces the search space to $2^{((L+1)/2)}$ and ensures that all sidelobes corresponding to odd shifts vanish, i.e., $C_k=0$ for all odd $k$. As a consequence, the total energy $E$ is reduced, which in turn increases the merit factor $F$. For sequence lengths $L\le 66$, only $22$ optimal sequences $S^*$ are also skew-symmetric~\cite{packebusch16}.

Since exhaustive search becomes computationally intractable for large values of $L$, restriction classes have been introduced~\cite{dimitrov22}. This approach enables a decomposition of the search space into multiple disjoint regions, which can then be explored in parallel. The search space can in turn be reduced to $2^{(L/2-p)}$ by fixing the first $p$ elements of the sequence. The first $p$ elements are determined using partitions (restriction classes)~\cite{dimitrov22} of length $g$ (number of summands) with minimal or normalized potentials. The next $k-p+1$ elements are free, while the last $k$ elements are determined using a skew-symmetry rule, as shown in Equation~\ref{eq:partitions}. This effectively assigns some elements in the autocorrelation function to small values, thereby reducing the total energy. To order the partitions, the authors in ~\cite{dimitrov22} suggested potentials and normalized potentials for each partition. This reduction of the search space, based on group theory, provides a more efficient approach for finding sequences with desired properties compared to considering all possible restriction classes.

\begin{equation}
\label{eq:partitions}
S(L)=\underbrace{s_1s_2 \cdots s_{p}}_{p}
\underbrace{s_{p+1}s_{p+2} \cdots s_{k-1}s_{k}s_{k+1}}_{k-p+1}
\underbrace{s_{k+2}s_{k+3} \cdots s_{L-1}s_{L}}_{k}    
\end{equation}

In this work, we address the challenge of efficiently exploring the exponentially large search space of binary sequences in the LABS problem while maintaining a strong balance between exploration and exploitation. We propose a hybrid framework that combines Thompson sampling–based online decision making with parallel self-avoiding walks to dynamically allocate computational resources across restriction classes. Each restriction class is treated as an arm in a multi-armed bandit setting, enabling adaptive focus toward regions of the search space that empirically yield higher merit factors, while still preserving sufficient exploration of under-sampled partitions. The resulting search procedure is further accelerated through GPU-parallel execution of independent walks, shared global posterior updates, and efficient neighborhood evaluation with linear-time flip operations and a Bloom filter for cycle prevention. In addition, we incorporate a two-stage optimization strategy that first explores constrained partitioned symmetric search spaces and subsequently refines high-quality candidates in an unconstrained setting, improving solution quality. Together, these components form a scalable and data-driven search strategy that leverages stochastic decision-making and high-performance parallel computation to significantly enhance the effectiveness of merit factor maximization.

Accordingly, the main contributions of this paper are as follows:
\begin{itemize}
\item We propose an online method for prioritizing regions of the search space in the low autocorrelation binary sequence problem, in which computational effort is dynamically allocated during the search process.
\item We develop a novel algorithm that incorporates the proposed prioritization strategy.
\item We report new best-known solutions for several sequence lengths.
\item We present the longest binary sequence with a merit factor exceeding $8.0$.
\end{itemize}


The remainder of this paper is organized as follows. We begin with a review of related work and background on low autocorrelation binary sequences. This is followed by the presentation of the proposed method for prioritizing regions of the search space and the corresponding algorithm. We then report the experimental results, including improved best-known solutions and a newly obtained sequence with a merit factor exceeding $8.0$. The paper concludes with a summary of findings and directions for future research.

\section*{Related work}

Finding optimal sequences $S^*$ is computationally feasible only for shorter lengths; nevertheless, several methods have been proposed in literature. In~\cite{mertens96}, the authors introduced a branch-and-bound algorithm that systematically explores the solution space. To reduce its size, they fixed the $m$ leftmost and rightmost elements based on symmetry rules. The lower bound of the energy function was obtained via relaxation, accounting for interactions between fixed and free elements. This approach enabled the computation of optimal sequences up to $L \le 44$, with an estimated time complexity of $O(1.85^L)$. The quality of the lower bound in~\cite{mertens96} depends on random assignments of the central $L-2m$ free elements. To address this,~\cite{prestwich07} introduced the concept of free products, where terms are considered only if both elements are unfixed. This idea was refined in~\cite{prestwich13} by incorporating additional interactions between fixed and free elements, reducing the complexity to $O(1.80^L)$. In~\cite{wiggenbrock10}, it was observed that conjugating a sequence element changes the energy sum by $\pm 4$, leading to a tighter lower bound. Finally, \cite{packebusch16} introduced a combined lower bound that further improved efficiency to $O(1.729^L)$. An exact lower bound was also derived; however, it was not used in practice due to its high computational cost. Using this method, optimal sequences were obtained up to $L \le 66$, with an estimated runtime of approximately $55$ days on a $248$-core system for $L = 66$. Additionally, optimal skew-symmetric sequences were reported for lengths up to $L \le 119$.

As an alternative, construction methods can be used to efficiently generate sequences of a given quality. In~\cite{hoholdt88}, it was shown that the merit factor of any rotated Legendre sequence satisfies $1.5 \le F \le 6.0$, attaining its maximum when the rotation parameter is approximately $r \approx \frac{1}{4}$. A Legendre sequence, defined for prime length $L$, whose elements are given by the Legendre symbol and take values $\pm1$ depending on quadratic residuosity (cf. quadratic residue modulo)~\cite{browein04}. Furthermore, \cite{browein04} demonstrated that sequences of arbitrary length with $F \ge 6.0$ can be systematically constructed. Specifically, by appending $\lfloor{\hat{t}L}\rfloor$ elements of a rotated Legendre sequence to the end, and tuning parameters within $0.20 \le r \le0.24$ and $0.055 \le \hat{t} \le 0.063$, one can achieve sequences with merit factor $F \approx 6.3421$. Further improvements using the steep descent algorithm and modified Jacobi sequences increased the merit factor to approximately $F \approx 6.44$~\cite{baden11}. A survey in~\cite{jedwab13} unified these results and identified sequence classes with asymptotic merit factors above $6.34$.

Despite these advances in construction methods, there remains a significant gap between asymptotic guarantees and best-known computationally obtained merit factors. The literature employed several stochastic approaches, including local search~\cite{farnane18,dimitrov25}, tabu search~\cite{halim08}, a combination of evolutionary algorithms with tabu search~\cite{gallardo09}, self-avoiding walks~\cite{boskovic17,boskovic24}, and search control using a priority queue~\cite{brest18,brest22}, among others. Local search methods in~\cite{boskovic17, dimitrov22} have produced sequences with a merit factor exceeding $8.0$ for $172 \le L \le 268$, and exceeding $7.0$ for $172 \le L \le 527$. In~\cite{brest22}, all skew-symmetric sequences with lengths in the interval $201 \leq L \leq 303$ achieving a merit factor of at least $F \ge 8.0$ were reported. Because exhaustive search becomes intractable for large values of $L$, the authors in~\cite{dimitrov22} introduced restriction classes, which allowed the search space to be divided into disjoint regions that can be explored in parallel. Analysis of a stochastic algorithm on short sequences enables the establishment of a predictive model for stopping conditions~\cite{herzog23}. Using this and parallel computation on graphics processing units~\cite{boskovic24}, the stochastic algorithm found some optimal skew-symmetric sequences with an estimated probability of $99\%$ for $171 \le L \le 223$, and skew-symmetric sequences with a merit factor greater than $9.0$ for lengths $225 \le L \le 247$.

Using noiseless simulations with up to 40 qubits, the authors in~\cite{shaydulin24} show that the quantum approximate optimization algorithm (QAOA) with fixed parameters achieves a time-to-solution scaling of $O(1.46^L)$  on the problem, which improves to $O(1.21^L)$  when combined with quantum minimum finding, outperforming the best classical heuristic. These results provide evidence that QAOA can serve as a useful algorithmic component for achieving quantum speedups in optimization problems in a fault-tolerant setting. In~\cite{sciorilli25}, the authors extend the Pauli Correlation Encoding (PCE) framework to the LABS problem. Their variational quantum solver achieves a time-to-solution scaling of approximately $O(1.33^L)$, and they project that quantum advantage could appear for instance sizes on the order of thousands.

In~\cite{zurek17}, GPU parallelization was shown to significantly speed up discrete optimization problems like LABS and the Golomb ruler, achieving about a $13\times$ speedup using steepest descent local search for sequences of length $L=201$. In~\cite{pietak19}, a hybrid memetic evolutionary multi-agent system (EMAS) uses GPUs to accelerate fitness evaluation, reaching a $46\times$ speedup for $L=128$ compared to CPU-only execution. The authors in~\cite{zurek22} used a deep neural network as a tabu support. They replaced the heuristic selection of the tabu parameter with an LSTM neural network that predicts values of the parameter for a given input sequence and its energy. In~\cite{zhang25} a framework for the massive parallelization of the memetic algorithm with tabu search was presented. In~\cite{cadavid25}, a quantum-enhanced memetic tabu search achieved the scaling of $O(1.24^L)$. In~\cite{boskovic24}, the authors presented a new stochastic algorithm for LABS, which organizes the search process as a series of parallel self-avoiding walks over sequences with skew-symmetry. The algorithm achieved a $387$-fold speedup compared to a sequential program, since the individual self-avoiding walks are trivially parallelizable. The algorithm achieved a time complexity of $O(1.34^L)$ and $O(1.15^L)$ for skew-symmetric sequences. In~\cite{psenicnik25}, the solver was improved to use partitions (restriction classes) and a two-step targeted refinement strategy was suggested to explore the whole unrestricted search space and mitigate constraints imposed by skew-symmetry and restriction classes.

\section*{Method}
Achieving high merit factors for long binary sequences requires careful selection of integer partitions, as exhaustively optimizing sequences for all partitions for a given length $p$ is computationally infeasible. Until now, potentials and normalized potentials of partitions have been used as heuristic methods to rank partitions~\cite{dimitrov22}, enabling the prioritization of promising regions of the search space. In this paper, we propose a novel method for dynamic allocation of search effort, which is modeled as an online decision problem. The proposed method dynamically allocates computational resources to partitions based on their observed performance, enabling efficient exploration and exploitation of the search space.

An online decision problem is a setting where a decision-maker (often called an agent) must make decisions sequentially over time without knowing future inputs. The agent only sees information available at the current time step and must act immediately. This process happens over discrete time stamps $t=1,2,3, \dots, T$, where at each time step an action is chosen based on past and current observations. After taking an action $a_t$, the agent receives a reward $r_t$. The effectiveness of online decision-making algorithms is commonly evaluated and compared using regret plots. The regret incurred at time $t$ is defined as the difference between the expected reward of the optimal action and the expected reward of the action chosen by the algorithm~\cite{russo18}. The objective is therefore to minimize cumulative regret over time.

We model the partition resource allocation problem as a multi-armed bandit problem~\cite{katehakis87}, where each partition corresponds to an arm. At each decision step, the algorithm selects partitions and allocates computational resources to them. Feedback is then received in the form of the best merit factor achieved after partitioning the sequence and performing a self-avoiding walk~\cite{boskovic24} on the partitioned sequence. Since we can only obtain an optimized merit factor of a partitioned sequence after performing the walk (and spending computation time), and because its value is random due to the stochastic nature of the search process, this problem involves a balance between exploration and exploitation. The goal is therefore to allocate resources to partitions, that are likely to yield high merit factors, while still exploring less-sampled partitions to avoid prematurely discarding potentially promising regions of the search space. To address this trade-off, we employ Thompson sampling~\cite{thompson33, thompson35}, which was chosen due to the low computational overhead compared to other reinforcement learning methods and strong empirical performance~\cite{agarwal10, chapelle11}.

Formally, let $K$ be the number of available partitions for a given sequence length. Each arm $k \in \{1, 2, \ldots, K\}$ corresponds to a partition and is associated with an unknown parameter $\theta_k$, representing the expected merit factor $F$ obtainable from that partition. We place a prior distribution $p(\theta_k)$ over each $\theta_k$ to encode our initial belief about the quality of partition $k$. A common choice for a prior distribution is the Beta distribution:
\begin{equation}
    \theta_k \sim Beta(\alpha_k,\beta_k),
\end{equation}
as it provides a convenient conjugate prior for bounded rewards in $[0,1]$ and enables efficient posterior updates. Here, $\alpha_k$ and $\beta_k$ correspond to the number of observed successes and failures, respectively, in a Bernoulli bandit framework. We initialize all arms with $\alpha_k = 1$ and $\beta_k = 1$, corresponding to a uniform prior.

However, since the merit factor $F$ is a continuous quantity, the reward $r_t$ at time $t$ cannot be effectively treated as a binary outcome. To address this, we employ an extension of Thompson sampling, proposed in~\cite{agrawal12}, which accommodates general reward distributions and thus enables modeling the full stochastic bandit problem. Specifically, for a reward $r_t \in [0,1]$, one can simulate a Bernoulli trial with success probability $r_t$ to map the reward to a binary outcome. An alternative, more direct approach allows updates with the actual reward value $r_t \in [0, 1]$ without performing a Bernoulli trial. This method, while not analyzed due to limitations~\cite{agrawal12}, is adopted here for efficiency. Under Fractional Thompson sampling, the parameters for arm $k$ are updated as:
\begin{equation}
(\alpha_k, \beta_k) \gets 
\begin{cases} 
(\alpha_k, \beta_k), & \text{if } a_t \neq k \\
(\alpha_k, \beta_k) + (r_t, 1 - r_t), & \text{if } a_t = k
\end{cases}
\end{equation}
Using fractional updates avoids the inefficiency of simulating Bernoulli trials for continuous rewards while preserving the core advantages of Thompson sampling. We scale the observed merit factor to $r_t \in [0, 1]$ for compatibility with the bandit framework, using the formula $r_t=clamp(F/7)$ based on the expected range of merit factors~\cite{psenicnik25}.

At each decision step $t$, a partition is chosen according to Thompson sampling. For each arm $k \in \{1, 2, \dots, K\}$, we draw a sample $\tilde{\theta}_k$ which represents a plausible estimate of the expected reward for partition $k$ under the current posterior distribution. The selected arm is then given by

\begin{equation}
    a_t = \arg\max_{k} \tilde{\theta}_k.
\end{equation}

Once a partition is selected, computational resources are allocated to it by performing a self-avoiding walk on the corresponding partitioned sequence. This gives a reward $r_t$, defined as the normalized merit factor obtained during the search. The parameters of the chosen arm are then updated using the fractional Thompson sampling rule described above. This process repeats until the stopping condition is achieved.

Based on the proposed approach, we introduce a new algorithm for generating binary sequences with high merit factors, called TS-SAW. The pseudo-code of the algorithm is presented in Algorithm~\ref{algo:tsaw}, while a flowchart illustrating the overall procedure is provided in Figure~\ref{fig:flowchart}. To reduce the time and memory complexity of the proposed method, and to enable efficient use of multi-threading on modern processors and graphical processing units, the $\mathtt{BEST\_NEIGHBOUR}$ function employs lightweight flip probing of skew-symmetric binary sequences with linear time and space complexity, as suggested in~\cite{dimitrov25}. This results in a total complexity of $O((L/2 - p)\cdot L) = O(L^2)$ for each call, enabling an efficient neighborhood evaluation mechanism requiring minimal memory. The neighborhood of a binary sequence consists of all sequences that differ from the original by a single element. Once all neighbors of the current pivot (current sequence in the self-avoiding walk) have been evaluated, the new pivot is selected as the unvisited neighbor with the minimum energy. In the same work~\cite{dimitrov25}, an in-memory flip algorithm for skew-symmetric binary sequences with linear time and space complexity was also proposed, and is adopted here. The function maintains skew-symmetry and flips the elements at indexes $i$ and $(L-q)-1$.

\input{ALG_TSAW}
\input{FIG_FLOWCHART}

To prevent cycling in the search process, the algorithm stores sequences' hashes, which limits the walk length $T_i$ due to memory constraints. Since high performance depends on a small memory footprint, so the data can fit in fast GPU shared memory and CPU cache, we use a Bloom filter~\cite{bloom70} in the suggested algorithm. This reduces memory usage and allows longer walks, while still providing a fast lookup for previously visited sequences, with a negligible chance of error.

Based on our previous research~\cite{boskovic24, psenicnik25}, graphical processing units offer several advantages for the LABS problem. Each walk is independent, allowing efficient parallelization with minimal synchronization overhead. In our implementation, each self-avoiding walk, as described in Algorithm~\ref{algo:tsaw}, is executed within a CUDA block with multiple threads, enabling multiple walks to run concurrently. This allows micro and macro parallelization of the algorithm, which improves efficiency and speed. To share information across walks and reduce entropy, a single global belief, represented by $\alpha_k$ and  $\beta_k$ parameters of each arm, is obtained and updated after each walk. Each walk can therefore be seen as an independent agent, sampling from the shared posterior and selecting a trajectory accordingly, which promotes exploration diversity. The global belief is then updated using the aggregated outcomes of all walks. While updates are delayed due to batching, shared observations accelerate learning and improve overall efficiency.

In~\cite{psenicnik25}, we showed that combining parallel stochastic search with targeted refinement is an effective strategy for solving extremely challenging combinatorial optimization problems such as LABS. Statistical tests further confirmed that this two-step method significantly outperforms single-step approaches. In this work, we adopt the same overall approach, but modify the transition between the two steps. Instead of filtering candidate solutions for the second step using a fixed merit factor threshold, we select $m$ best candidates based on the second-step capacity. These candidates are chosen by ranking all parallel self-avoiding walks according to their merit factor and selecting the top subset using the quickselect algorithm, which operates in expected linear time. Although the second step itself remains unchanged from the original paper, its pseudocode is provided in Algorithm~\ref{algo:pq} for completeness. The purpose of the second step is to explore the full search space without imposing the skew-symmetry and partition constraints used in the first step in the hopes of improving the solution quality further. For clarity, the overall approach is depicted in Figure~\ref{fig:flow}. The $\mathtt{SEQUENCE\_FLIP}$ function flips the $i$-th element of a sequence in linear time and space complexity. It differs from the $\mathtt{SEQUENCE\_FLIP\_SKEW}$ function in that it modifies only a single element and does not preserve skew-symmetry. The $\mathtt{MAKE\_ROTATIONS}$ function generates multiple variants of a given pivoting sequence by rotating it left and right by up to $T_r$ positions, where each rotation shifts elements cyclically while preserving the sequence length:
\begin{equation}
    s'[i]=s[(i+r)\bmod \, L].
\end{equation}

Because these rotations only rearrange the positions of elements without changing their values, the resulting sequences typically have similar energy (or merit factor) to the original sequence. Each rotated sequence is then inserted into the priority queue for further evaluation. Unlike transformations used in the first step, this function does not preserve skew-symmetry or the restriction class as well, which allows the algorithm to explore a broader region of the search space.
This design enables continuous and efficient utilization of computational resources, since both steps can run in parallel, allowing the system to maintain high throughput while minimizing idle CPU and GPU time.

\input{FIG_FLOW}
\input{ALG_DFS}

To extend the improvements achieved for odd-length binary sequences to the even-length case, sequence operators~\cite{dimitrov22} can be systematically employed. These operators act by appending or removing elements at either end of a given sequence, thereby generating new candidate sequences of modified length while possibly preserving key properties. In particular, starting from optimized odd-length sequences, such transformations enable the construction of both odd- and even-length variants with comparable merit factor $F$. The resulting sequences are subsequently reintroduced into the second optimization step, where they serve as promising initial candidates for further refinement.

In summary, the proposed method integrates adaptive resource allocation with large-scale parallel stochastic search to efficiently navigate the combinatorial space of binary sequences. By combining Thompson sampling–based partition selection, GPU-accelerated self-avoiding walks, and a two-step refinement strategy, the approach enables both effective exploration of constrained subspaces and thorough exploitation of promising candidates in the unconstrained domain. The inclusion of sequence operators further extends the applicability of the method to sequences of arbitrary length, ensuring a unified framework for both odd- and even-length cases. This design provides a practical and scalable foundation for high-performance optimization in the LABS problem.

\section*{Results}
We implemented the proposed method and Algorithm~\ref{algo:tsaw} in the C++ programming language using CUDA Toolkit version 13.2. Algorithm~\ref{algo:pq} was implemented in the Rust programming language (version 1.95). All experiments were conducted on the Vega grid computing environment, Slovenia’s petascale supercomputer, using \textit{NVIDIA A100-SXM4-40GB} GPUs. Unless stated otherwise, all experimental parameters were adopted from our previous study~\cite{psenicnik25}, in which they were systematically tuned.

To evaluate the partitions (restriction classes) for required values of $p$, we employed the algorithm presented in~\cite{eppstein15}. The potentials and normalized potentials were then computed as described in~\cite{dimitrov22}; by partitioning a sequence of sufficient length and setting the middle $k-p+1$ free elements to the neutral element $0$, as shown in Equation~\ref{eq:partitions}. In addition, the number of summands $g$ for each partition was recorded. As a note, the potential and normalized potential are invariant of the sequence length $L$. For transparency and reproducibility, top partitions ordered by increasing normalized potential for $p=81$ and $q=7$ together with their corresponding potential and normalized potential values are reported in Table~\ref{tab:toppar}.

\input{TAB_TOPPAR}

To improve the best binary sequences reported in the literature, we applied the proposed approach to all odd-length binary sequences in the range $450 \le L \le 527$, as well as to $L=573$, using $100,\!000$ iterations. An iteration corresponds to a single execution across all blocks of Algorithm~\ref{algo:tsaw}, in which block-size parallel self-avoiding walks are performed. Then the best $m$ sequences are selected and optimized with the second step. In this work, we set $m=15$, mainly considering the computational constraints of a single node in the computing grid and the runtime of an iteration of the first step, without performing any parameter tuning. To extend the improvements to even lengths, the resulting partitioned skew-symmetric sequences were transformed to even lengths using sequence operators and then reintroduced to the second step as show in Figure~\ref{fig:flow}. Overall, this procedure led to improvements for $35$ sequences. The resulting improved binary sequences and their corresponding merit factors, in comparison with the previously best reported results in literature~\cite{psenicnik25,dimitrov22}, are presented in  Figure~\ref{fig:records} and summarized in Table~\ref{tab:records}. It should also be noted that in Figure~\ref{fig:records}, the x-axis (sequence length interval) is not continuous, as only selected discrete lengths are included for which improvements were obtained. In Table~\ref{tab:records}, each hexadecimal digit is mapped to a $4$-bit binary representation (i.e., $0=0000$, $1=0001$, $\dots$, $F=1111$). To obtain the correct sequence length, leading zero-padding introduced by this encoding must be removed. Finally, the binary sequences are reconstructed by mapping each binary digit to its corresponding bipolar value, specifically by converting every $0 \rightarrow -1$.

\input{FIG_RECORDS}

The largest improvement was achieved for $L=451$, where the previously best-known merit factor was increased by $0.8248$. The resulting sequence attains a merit factor of $F=8.0555$, making it the longest sequence reported in the literature with $F \ge 8.0$. Previously, the longest known sequence satisfying this condition was the sequence of length $L=309$ reported in~\cite{dimitrov22}. In addition, $L=573$ is the longest sequence in the literature with a merit factor above $7.0$, for which we achieved an improvement of $0.2564$. For $L=518$, the merit factor was also improved; notably, this was the only sequence in the considered interval that was not improved by the solver in~\cite{psenicnik25}. Across all improved sequences, the mean increase in merit factor was $0.1967$. These results demonstrate the effectiveness of the proposed method in discovering high-quality binary sequences with large merit factors, particularly for longer sequence lengths. The consistent improvements observed across a broad range of lengths suggest that prioritizing promising regions of the search space provides a robust, practical, and scalable foundation for high-performance optimization of the LABS problem.

To show how Thompson sampling affects partition selection, we conducted an additional experiment with sequence length $L=461$, limiting the proposed optimizer to $80,\!000$ iterations. We selected partition parameters $p=67$ and $g=5$ based on the sequence length. For that pair, the number of unique partitions is $8056$. The goal of this experiment was to analyze how the probabilistic selection mechanism influences exploration and exploitation of the partition space over time, and whether it leads to a bias toward higher-quality regions of the search space. In particular, we tracked the frequency with which individual partitions were sampled and their $(\alpha_k, \beta_k)$ values during optimization. Table~\ref{tab:partitions} presents the most frequently sampled partitions, along with the number of times each partition was selected, its ranking according to normalized potential, and the corresponding normalized potential value. The table includes the three most frequently sampled partitions and the three highest-ranked partitions according to normalized potential. It is evident that a clear mismatch occurs between the number of times a partition was selected and its rank when ordered by the normalized potential. This result indicates that partitions with the highest normalized potential are not necessarily the most effective during optimization. Thompson sampling progressively favors partitions that produce better empirical results, even when their normalized potential ranking is lower. Consequently, the sampling process is guided not only by the heuristic estimate (in case not every partition of a given pair is used), but also by the observed optimization performance of each partition. To provide further insight, we analyzed the probability density functions (PDFs) corresponding to the best-ranked partition, the most frequently sampled partition, and a partition with rank 4, representing average performance. As a note, a PDF describes the distribution of a continuous random variable by assigning relative likelihoods to its possible values. Probabilities are obtained by integrating the PDF over an interval, with the total area under the function equal to one.

\input{TAB_PARTITIONS}
\input{FIG_PARTITIONS_PDF}

During the early stages of the search, when only a limited number of observations are available, the posterior distributions exhibit substantial overlap, indicating considerable uncertainty regarding the relative quality of the selected partitions. As additional observations are collected, the distributions become increasingly concentrated and better separated, reflecting reduced uncertainty and more accurate estimates of expected rewards. By iteration $7200$, the posterior distribution corresponding to the most frequently sampled partition is clearly centered at the highest expected value, with minimal overlap with the other distributions, indicating that it is the most promising candidate. Initially, all three partitions are sampled at similar rates due to the high level of uncertainty. As the search progresses and confidence in the reward estimates increases, the sampling process gradually shifts toward the partition with the highest estimated reward, illustrating the balance between exploration and exploitation achieved by Thompson sampling.

In general, the results show that the proposed framework gradually shifts from uniform exploration toward partitions that yield higher observed rewards. In early iterations, a broad range of partitions is sampled, while later stages focus on a smaller subset of consistently high-performing partitions. This effect is most pronounced in regions associated with high-quality sequences, while less promising partitions are increasingly suppressed, reducing wasted computation. Overall, this provides a good balance between exploration and exploitation, concentrating search effort on promising regions while maintaining diversity in the early phase.

Despite the improvements reported in this work, the results also confirm the increasing difficulty of achieving substantial merit-factor gains as the sequence length grows. For larger values of $L$, improvements become smaller and less frequent, consistent with the increasingly rugged optimization landscape of the LABS problem. Nevertheless, obtaining improvements for long sequences, including $L=573$, demonstrates the scalability of the proposed framework and its suitability for large-scale parallel computing environments. These results further indicate that partition-guided large-scale stochastic optimization remains a promising direction for advancing the state-of-the-art in the LABS problem.

\input{TAB_RECORDS}

\section*{Conclusion}
In this work, we addressed the problem of low autocorrelation binary sequences (LABS). The problem is characterized by an exponentially growing search space and a highly rugged landscape. We proposed a hybrid framework that combines Thompson sampling–based online decision-making with parallel self-avoiding walks to allocate dynamically computational resources across restriction classes (partitions). The main idea of the proposed approach is to model the partitions as a stochastic decision-making problem in Thompson sampling. This allows the framework to dynamically allocate computational resources to promising regions of the search space. The resulting method effectively balanced exploration and exploitation while maintaining scalability through parallel execution.

We proposed an algorithm based on the above method to improve the results. Experimental results on longer binary sequences show improvements over previously best-known solutions across a wide range of sequence lengths. We report new best-known sequences for multiple lengths in the range $450 \le L \le 527$ and $L=573$, including the longest sequence with a merit factor exceeding $8.0$, for $L=451$. We also showed that Thompson sampling effectively shifts search effort toward partitions that yield higher merit factors. This allowed us to empirically measure partition quality instead of relying on heuristics. Overall, the results confirm that integrating probabilistic decision-making with large-scale parallel stochastic search is an effective strategy for LABS optimization. Although improvements become more challenging as sequence length increases, the proposed approach remains effective, even in higher dimensions, and continues to yield state-of-the-art results. 

We plan to further investigate the proposed online decision-based prioritization of search regions by applying it to other difficult combinatorial optimization problems beyond LABS. A key direction is to study problem classes in which the search space can be meaningfully divided into structured subregions, so that computational effort can be adjusted dynamically according to their observed performance during the search. We also intend to explore how the design insights obtained from LABS optimization can be adapted for the construction of spreading codes in satellite communication systems. Since low autocorrelation binary sequences are closely related to spreading code design used in GNSS and similar applications. As a conclusion, we believe that the proposed framework opens a pathway toward adaptive, data-driven search strategies for a wider class of combinatorial optimization problems.

\bibliographystyle{unsrt}
\bibliography{references}

\end{document}

%% file: FIG_SEARCH_SPACE.tex
\begin{figure}[htb]
    \centering
    \includegraphics[width=.85\linewidth]{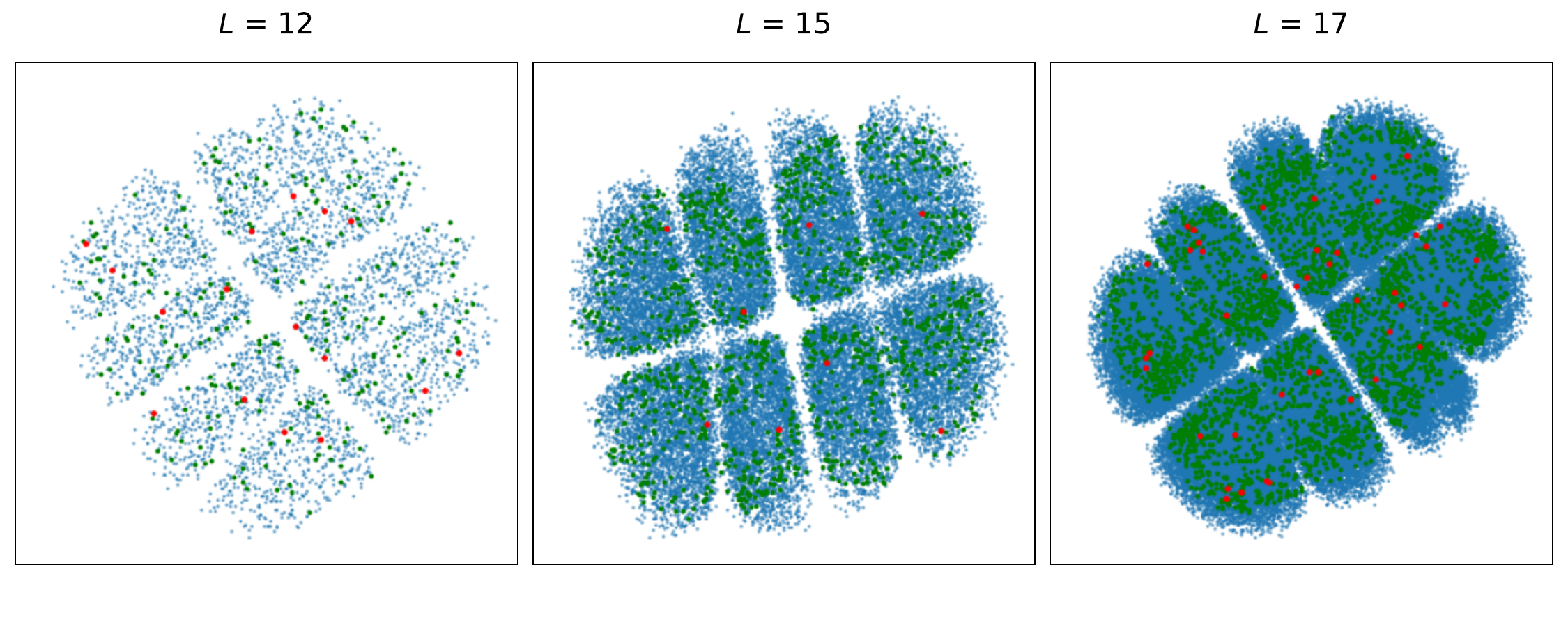}
    \caption{Two-dimensional projection of the search space for sequence lengths $L=12,\,15,\,17$ using the dimensionality reduction method UMAP~\cite{healy24}. Red points represent global minima, green points represent local minima, and blue points correspond to all other sequences.}
    \label{fig:search_space}
\end{figure}

%% file: ALG_TSAW.tex
\begin{algorithm}[tbh]
\caption{Pseudocode of the proposed TS-SAW algorithm for generating binary sequences with high merit factors using search space prioritization.}\label{algo:tsaw}
\begin{small}
\begin{algorithmic}[1]
    \State $\mathbb{BF} \gets empty $ \Comment{Create an empty Bloom Filter}
    \For{$k = 1, \dots, K$}
        \State $\hat{\theta}_k \sim \text{Beta}(\alpha_k, \beta_k)$
    \EndFor
    \State $a_t = \arg\max_{k} \tilde{\theta}_k $ \Comment{Select a promising search region}
    \State $seq \gets \Call{init\_partitioned\_sequence}{a_t}$
    \State $seq_{best} \gets seq$
    \State $\mathbb{BF}.\Call{insert}{seq}$
    \State $e \gets \Call{energy}{seq}$
    \State $e_{best} \gets e$
    \State $it \gets 0$
    \While{$it < T_i $} \Comment{Self-avoiding walk of length $T_i$}
    \State $it \gets it + 1$
            \State $i, delta \gets \Call{best\_neighbour}{seq}$
            \If {$i = -1$}
        \State \textbf{break}
    \EndIf
            \State $\Call{sequence\_flip\_skew}{seq, i}$
    \State $\mathbb{BF}.\Call{insert}{seq}$
            \State $e \gets e + delta$
            \If {$e < e_{best}$}
        \State $seq_{best} \gets seq$
        \State $e_{best} \gets e$
    \EndIf
    \EndWhile
    \State $r_t \gets \Call{calculate\_reward}{seq_{best}}$
    \State $(\alpha_{a_t}, \beta_{a_t}) \gets (\alpha_{a_t}, \beta_{a_t}) + (r_t, 1 - r_t)$ \Comment{Update beliefs about the selected search region}
\end{algorithmic}
\end{small}
\end{algorithm}

%% file: FIG_FLOWCHART.tex
\begin{figure}
    \centering
    \includegraphics[width=1\linewidth]{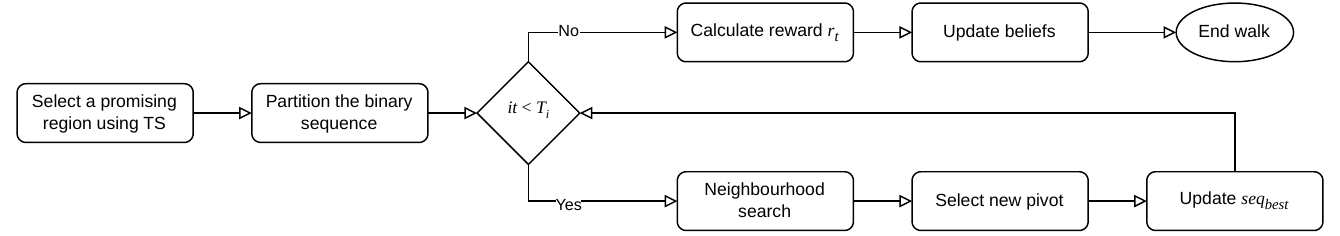}
    \caption{Flowchart illustrating the main steps of the proposed TS-SAW algorithm for generating binary sequences with high merit factors.}
    \label{fig:flowchart}
\end{figure}

%% file: FIG_FLOW.tex
\begin{figure}
    \centering
    \includegraphics[width=.85\linewidth]{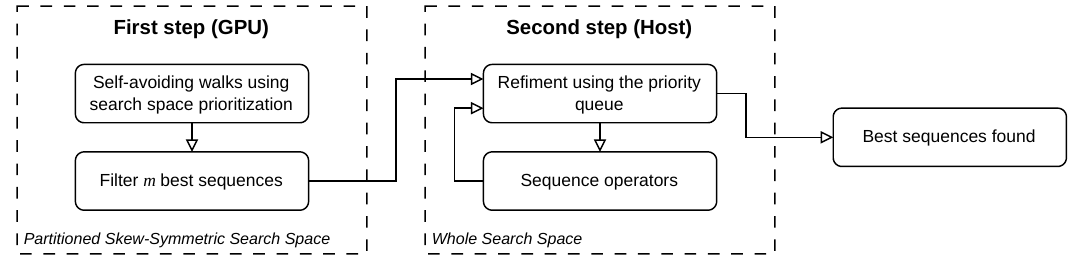}
    \caption{Two-step optimization pipeline combining parallel stochastic search and unconstrained refinement. The top $m$ candidates are selected via quickselect and passed to a second phase for further improvement.}
    \label{fig:flow}
\end{figure}

%% file: ALG_DFS.tex
\begin{algorithm}[t!bh]
\caption{Second step using a priority queue}\label{algo:pq}
\begin{small}
\begin{algorithmic}[1]
\Require{$candidate$} \Comment{Candidate from the first phase (SAW)}
\Ensure{$seq_{best}$} \Comment{Binary sequence with the best merit factor found}
    \State $\mathbb{PQ}, \mathbb{H} \gets empty, empty\ $ \Comment{Empty priority queue and hash set}
    \State $seq_{best} \gets candidate$
    \State $e_{best} \gets \Call{energy}{candidate}$
        \State $\mathbb{PQ}.\Call{push}{candidate}$
        \State $u \gets 0$
        \While{$u < T_u$}
        \State $u \gets u + 1$
        \State $seq_{current} \gets \mathbb{PQ}.\Call{pop}$
                \For{$\textbf{each}\ i \in 0..L$} \Comment{Neighborhood search}
            \State $hash \gets \Call{hash\_flip}{i}$
            \If {$\mathbb{H}.\Call{contains}{hash}$}
                \State \textbf{continue}
            \EndIf
                        \State $seq_{current} \gets \Call{sequence\_flip}{seq_{current}, i}$
                        \State $\mathbb{PQ}.\Call{push}{seq_{current}}$
            \State $\mathbb{H}.\Call{insert}{hash}$
            \State $e_{current} \gets \Call{energy}{seq_{current}}$
                        \If {$e_{current} < e_{best}$}
                \State $seq_{best} \gets seq_{current}$ \Comment{New best sequence found}
                \State $e_{best} \gets e_{current}$
                \State $u \gets 0$ \Comment{Reset $u$ to support dynamic depth}
            \EndIf
            \State $\Call{make\_rotations}{seq_{current}, \mathbb{PQ}, \mathbb{H}, T_r}$
            \State $seq_{current} \gets \Call{sequence\_flip}{seq_{current}, i}$
        \EndFor
    \EndWhile
        \State \Return $seq_{best}$ \Comment{Best sequence found}
\end{algorithmic}
\end{small}
\end{algorithm}

%% file: TAB_TOPPAR.tex
\begin{table}
    \centering
    \begin{tabular}{|c|c|c|c|c|c|}
        \hline
        \textbf{Rank} & \textbf{Partition} & $p$ & $q$ & \textbf{Potential} & \textbf{Normalized potential} \\
        \hline
        $1$ & $30\ \ 11\ \ 9\ \ 6\ \ 6\ \ 4$ & $81$ & $7$ & $17,\!745$ & $5,\!073$ \\
        $2$ & $30\ \ 13\ \ 12\ \ 8\ \ 6\ \ 6\ \ 6$ & $81$ & $7$ & $17,\!513$ & $5,\!081$ \\
        $3$ & $30\ \ 15\ \ 12\ \ 8\ \ 6\ \ 6\ \ 4$ & $81$ & $7$ & $17,\!857$ & $5,\!089$ \\
        $4$ & $32\ \ 13\ \ 12\ \ 8\ \ 6\ \ 6\ \ 4$ & $81$ & $7$ & $17,\!961$ & $5,\!121$ \\
        $5$ & $32\ \ 14\ \ 11\ \ 8\ \ 6\ \ 6\ \ 4$ & $81$ & $7$ & $17,\!921$ & $5,\!129$ \\
        \hline
    \end{tabular}
    \caption{Partitions for $p=81$ and $q=7$, ordered by increasing normalized potential, together with their corresponding potential and normalized potential values.}
    \label{tab:toppar}
\end{table}

%% file: FIG_RECORDS.tex
\begin{figure}
    \centering
    \includegraphics[width=.85\linewidth]{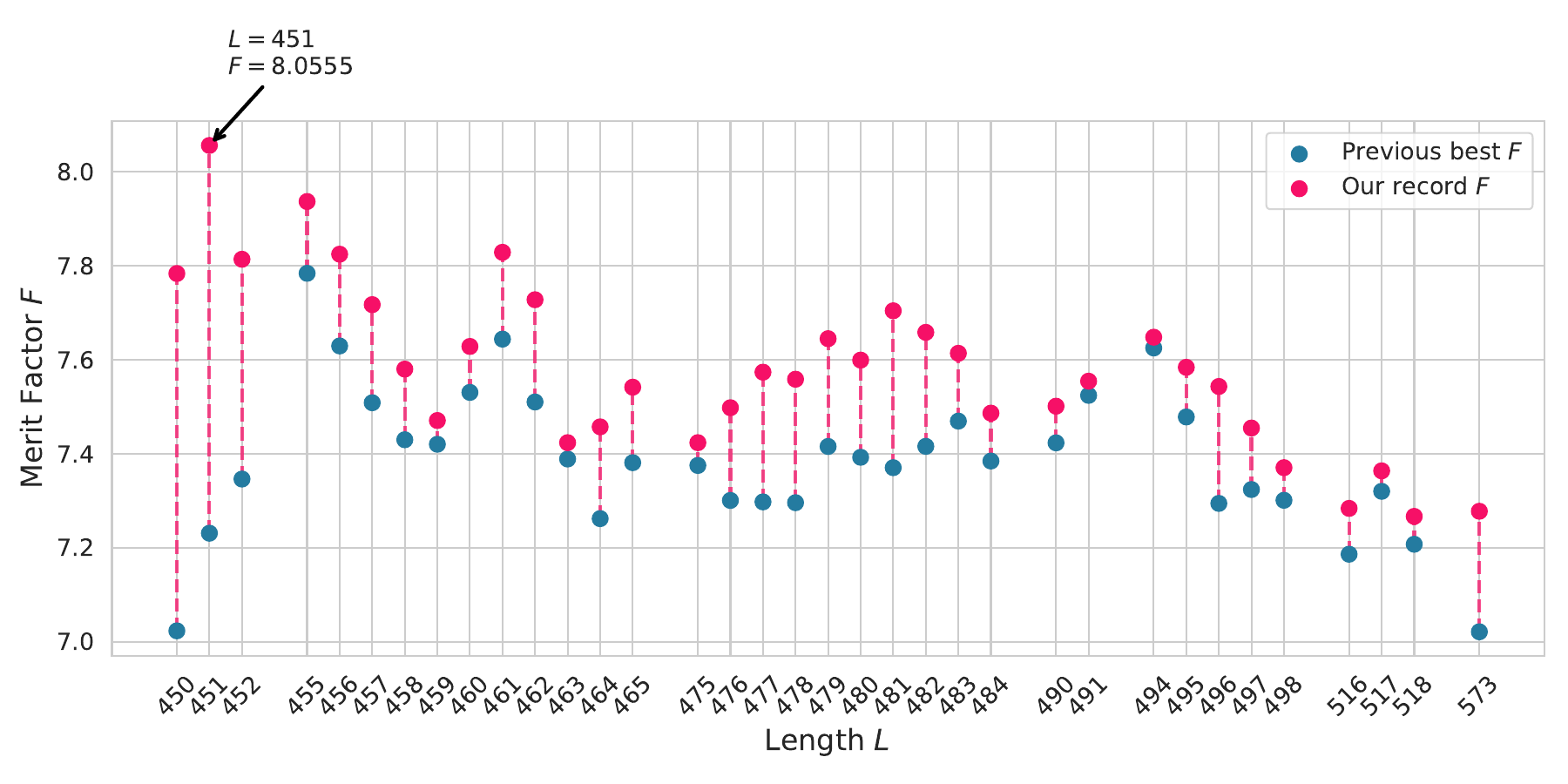}
    \caption{Record merit factors achieved by the improved sequences, together with the previously best-known merit factors reported in the literature~\cite{psenicnik25,dimitrov22}, for sequence lengths in the interval $450 \le L \le 527$ and for $L=573$.}
    \label{fig:records}
\end{figure}

%% file: TAB_PARTITIONS.tex
\begin{table}
    \centering
    \begin{tabular}{|c|c|c|c|c|c|}
        \hline
        \textbf{Partition} & \textbf{Sample rank} & \textbf{Pulls} & \textbf{Rank} & \textbf{Norm. potential} \\
        \hline
        $28\ \ 13\ \ 12\ \ 8\ \ 6$ & $1$ & $3,\!738,\!900$ & $205$ & $5,\!131$ \\
        $28\ \ 13\ \ 12\ \ 9\ \ 5$ & $2$ & $3,\!712,\!249$ & $274$ & $5,\!203$ \\
        $27\ \ 14\ \ 11\ \ 9\ \ 6$ & $3$ & $3,\!173,\!369$ & $309$ & $5,\!227$ \\
        $28\ \ 16\ \ 11\ \ 8\ \ 4$ & $334$ & $3,\!861$ & $1$ & $4,\!707$ \\
        $28\ \ 15\ \ 12\ \ 8\ \ 4$ & $49$ & $20,\!602$ & $2$ & $4,\!715$ \\
        $29\ \ 15\ \ 11\ \ 8\ \ 4$ & $16$ & $76,\!072$ & $3$ & $4,\!723$ \\
        \hline
    \end{tabular}
    \caption{Most frequently sampled and top ranked partitions by normalized potential for $L=461$ after $80,\!000$ iterations for pair $p=67$ and $g=5$. The table shows sampling rank, sampling frequency, normalized potential rank, and normalized potential value, comparing empirical selection with heuristic ranking under Thompson sampling.}
    \label{tab:partitions}
\end{table}

%% file: FIG_PARTITIONS_PDF.tex
\begin{figure}
    \centering
    \includegraphics[width=1\linewidth]{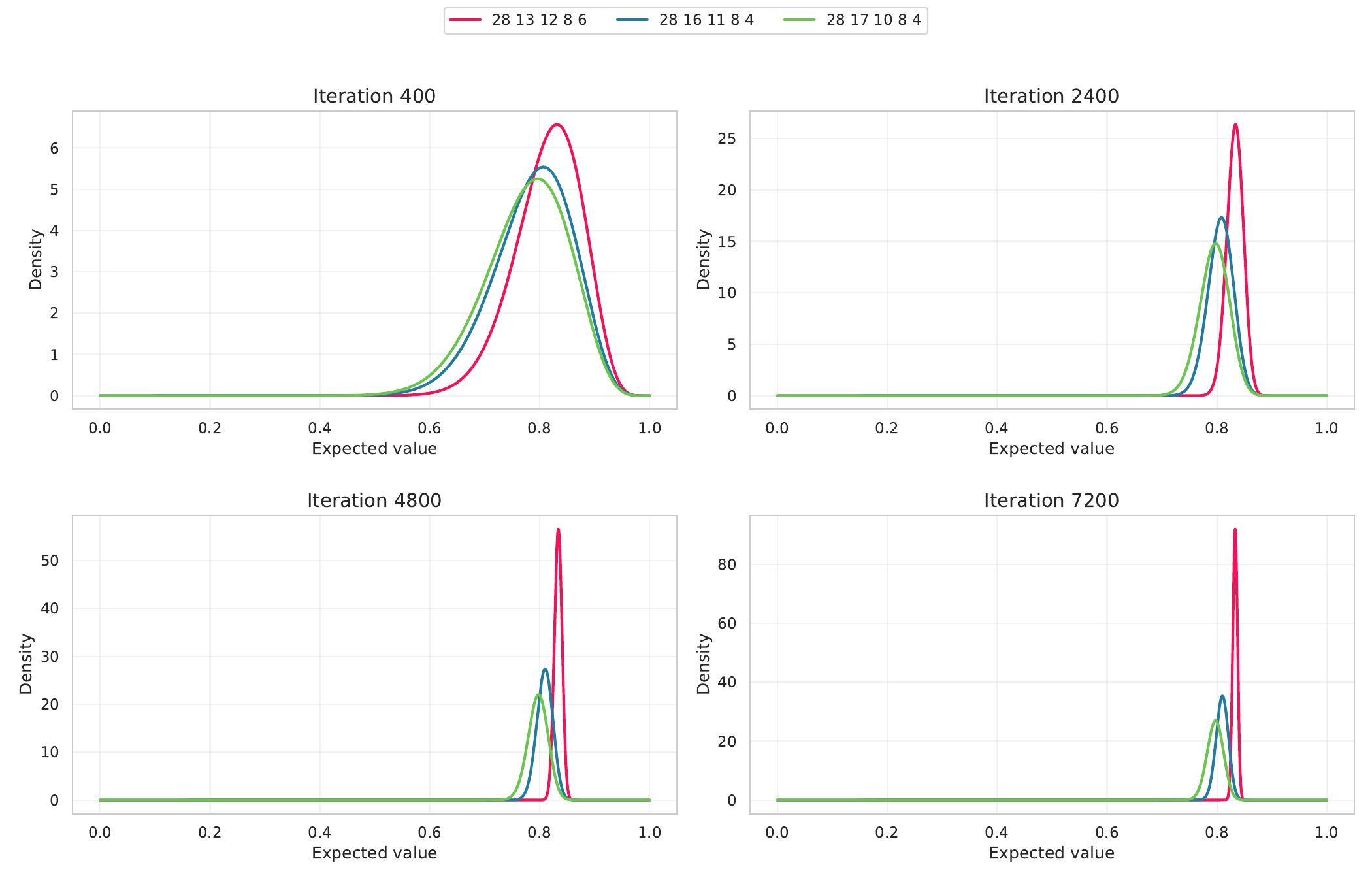}
    \caption{Probability density functions at different iterations for selected partitions for $L=461$ and pair $p=67$ and $g=5$.}
    \label{fig:pdf}
\end{figure}

%% file: TAB_RECORDS.tex

\newcolumntype{P}[1]{>{\centering\arraybackslash}p{#1}}

\begin{longtable}{|c|P{2cm}|c|p{\dimexpr\textwidth-6cm\relax}|}
\caption{Improved binary sequences and their merit factor values in the interval $450 \le L \le 527$ and for $L=573$, and the previously best-known merit factor values reported in~\cite{psenicnik25, dimitrov22}.}\label{tab:records} \\
\hline
\textbf{$L$} & \textbf{Previous $F$ ~\cite{psenicnik25, dimitrov22}} & \textbf{Our $F$}& \textbf{Sequence} \\
\hline
\endfirsthead

\hline
\textbf{$L$} & \textbf{Previous $F$ ~\cite{psenicnik25, dimitrov22}} & \textbf{Our $F$}& \textbf{Sequence} \\
\hline
\endhead

\hline
\multicolumn{4}{r}{Continued on the next page} \\
\endfoot

\hline
\endlastfoot

$450$ & $7.0230$ & {$\mathbf{7.7831}$} & 3\allowbreak f\allowbreak f\allowbreak f\allowbreak f\allowbreak f\allowbreak f\allowbreak 8\allowbreak 0\allowbreak 0\allowbreak 3\allowbreak f\allowbreak f\allowbreak 0\allowbreak 3\allowbreak c\allowbreak 3\allowbreak 8\allowbreak 3\allowbreak 1\allowbreak c\allowbreak e\allowbreak 6\allowbreak 7\allowbreak 3\allowbreak 0\allowbreak c\allowbreak 1\allowbreak f\allowbreak e\allowbreak 0\allowbreak f\allowbreak 1\allowbreak 8\allowbreak 7\allowbreak c\allowbreak 3\allowbreak e\allowbreak 1\allowbreak 3\allowbreak 3\allowbreak 3\allowbreak 9\allowbreak 3\allowbreak e\allowbreak 8\allowbreak 5\allowbreak c\allowbreak e\allowbreak 7\allowbreak 9\allowbreak 8\allowbreak b\allowbreak 1\allowbreak c\allowbreak 9\allowbreak b\allowbreak 9\allowbreak c\allowbreak 9\allowbreak 3\allowbreak d\allowbreak 9\allowbreak a\allowbreak 6\allowbreak c\allowbreak 8\allowbreak 5\allowbreak e\allowbreak b\allowbreak 1\allowbreak b\allowbreak 3\allowbreak 3\allowbreak 1\allowbreak 6\allowbreak b\allowbreak 4\allowbreak a\allowbreak 5\allowbreak 9\allowbreak 2\allowbreak d\allowbreak 6\allowbreak a\allowbreak 9\allowbreak 4\allowbreak d\allowbreak 3\allowbreak 2\allowbreak 6\allowbreak 6\allowbreak c\allowbreak 9\allowbreak 3\allowbreak 5\allowbreak b\allowbreak 4\allowbreak b\allowbreak 5\allowbreak 2\allowbreak a\allowbreak b\allowbreak 5\allowbreak 5\allowbreak 5\allowbreak a\allowbreak a\allowbreak a\allowbreak a\allowbreak a\allowbreak a\allowbreak a \\
\end{longtable}

%% file: references.bib
@article{thompson33,
 ISSN = {00063444},
 URL = {http://www.jstor.org/stable/2332286},
 author = {William R. Thompson},
 journal = {Biometrika},
 number = {3/4},
 pages = {285--294},
 publisher = {[Oxford University Press, Biometrika Trust]},
 title = {On the Likelihood that One Unknown Probability Exceeds Another in View of the Evidence of Two Samples},
 urldate = {2025-11-12},
 volume = {25},
 year = {1933}
}

@article{thompson35,
  title={On the theory of apportionment},
  author={Thompson, William R},
  journal={American Journal of Mathematics},
  volume={57},
  number={2},
  pages={450--456},
  year={1935},
  publisher={JSTOR}
}

@article{russo18,
  title={A tutorial on thompson sampling},
  author={Russo, Daniel J and Van Roy, Benjamin and Kazerouni, Abbas and Osband, Ian and Wen, Zheng and others},
  journal={Foundations and Trends (R) in Machine Learning},
  volume={11},
  number={1},
  pages={1--96},
  year={2018},
  publisher={Now Publishers, Inc.}
}

@article{dimitrov22,
  title={New classes of binary sequences with high merit factor},
  author={Dimitrov, Miroslav},
  journal={arXiv preprint arXiv:2206.12070},
  year={2022}
}

@article{katehakis87,
  author = {Katehakis, Michael N. and Veinott Jr., Arthur F.},
  title = {The Multi-Armed Bandit Problem: Decomposition and Computation},
  journal = {Mathematics of Operations Research},
  volume = {12},
  number = {2},
  pages = {262--268},
  year = {1987},
  doi = {10.1287/moor.12.2.262}
}

@article{boskovic24,
  title={Parallel self-avoiding walks for a low-autocorrelation binary sequences problem},
  author={Bo{\v{s}}kovi{\'c}, Borko and Herzog, Jana and Brest, Janez},
  journal={Journal of Computational Science},
  volume={77},
  pages={102260},
  year={2024},
  publisher={Elsevier}
}

@article{chapelle11,
  title={An empirical evaluation of thompson sampling},
  author={Chapelle, Olivier and Li, Lihong},
  journal={Advances in neural information processing systems},
  volume={24},
  year={2011}
}

@article{agarwal10,
  title={{'A modern Bayesian look at the multi-armed bandit' by Steven L. Scott: Discussion.}},
  author={Agarwal, Deepak K},
  journal={Applied Stochastic Models in Business \& Industry},
  volume={26},
  number={6},
  year={2010}
}

@inproceedings{agrawal12,
  title={Analysis of thompson sampling for the multi-armed bandit problem},
  author={Agrawal, Shipra and Goyal, Navin},
  booktitle={Conference on learning theory},
  pages={39--1},
  year={2012},
  organization={JMLR Workshop and Conference Proceedings}
}

@article{psenicnik25,
title = {Dual-step optimization for binary sequences with high merit factors},
journal = {Digital Signal Processing},
volume = {165},
pages = {105316},
year = {2025},
issn = {1051-2004},
doi = {https://doi.org/10.1016/j.dsp.2025.105316},
url = {https://www.sciencedirect.com/science/article/pii/S1051200425003380},
author = {Blaž Pšeničnik and Rene Mlinarič and Janez Brest and Borko Bošković}
}

@article{dimitrov25,
title = {On the skew-symmetric binary sequences and the merit factor problem},
journal = {Digital Signal Processing},
volume = {156},
pages = {104793},
year = {2025},
issn = {1051-2004},
doi = {https://doi.org/10.1016/j.dsp.2024.104793},
url = {https://www.sciencedirect.com/science/article/pii/S1051200424004184},
author = {Miroslav Dimitrov}
}

@article{bloom70,
  title={Space/time trade-offs in hash coding with allowable errors},
  author={Bloom, Burton H},
  journal={Communications of the ACM},
  volume={13},
  number={7},
  pages={422--426},
  year={1970},
  publisher={ACM New York, NY, USA}
}

@article{golay72,
  author={Golay, M.},
  journal={IEEE Transactions on Information Theory}, 
  title={A class of finite binary sequences with alternate auto-correlation values equal to zero (Corresp.)}, 
  year={1972},
  volume={18},
  number={3},
  pages={449-450},
  doi={10.1109/TIT.1972.1054797}
}

@article{littlewood66,
author = {Littlewood, J. E.},
title = {On Polynomials $\sum^n \pm z^m, \sum^ne^{a_mi}z^m,z=e^{0i}$},
journal = {Journal of the London Mathematical Society},
volume = {s1-41},
number = {1},
pages = {367-376},
doi = {10.1112/jlms/s1-41.1.367},
year = {1966}
}

@inproceedings{jedwab04,
  title={A survey of the merit factor problem for binary sequences},
  author={Jedwab, Jonathan},
  booktitle={International Conference on Sequences and Their Applications},
  pages={30--55},
  year={2004},
  organization={Springer}
}

@article{bernasconi87,
	Author = {J. Bernasconi},
	Journal = {J. Physsique},
	Month = {4},
	Pages = {559--567},
	Title = {Low autocorrelation binary sequences: statistical mechanics and configuration space analysis},
	Volume = {48},
	Year = {1987}
}

@article{katz24,
  title={Moments of autocorrelation demerit factors of binary sequences},
  author={Katz, Daniel J and Ramirez, Miriam E},
  journal={Designs, Codes and Cryptography},
  pages={1--45},
  year={2024},
  publisher={Springer}
}

@article{shapiro68,
  title = {Fourth Test of General Relativity: Preliminary Results},
  author = {Shapiro, Irwin I. and Pettengill, Gordon H. and Ash, Michael E. and Stone, Melvin L. and Smith, William B. and Ingalls, Richard P. and Brockelman, Richard A.},
  journal = {Phys. Rev. Lett.},
  volume = {20},
  issue = {22},
  pages = {1265--1269},
  numpages = {0},
  year = {1968},
  month = {5},
  publisher = {American Physical Society},
  doi = {10.1103/PhysRevLett.20.1265},
}

@inproceedings{yang24a,
  title={Large-Scale GNSS Spreading Code Optimization},
  author={Yang, Alan and Mina, Tara and Boyd, Stephen and Gao, Grace},
  booktitle={Proceedings of the 37th International Technical Meeting of the Satellite Division of The Institute of Navigation (ION GNSS+ 2024)},
  pages={948--957},
  year={2024}
}

@article{yang24b,
  title={Spreading code optimization for low-earth orbit satellites via mixed-integer convex programming},
  author={Yang, Alan and Mina, Tara and Gao, Grace},
  journal={EURASIP Journal on Advances in Signal Processing},
  volume={2024},
  number={1},
  pages={67},
  year={2024},
  publisher={Springer}
}

@article{sciorilli25,
  title={A competitive NISQ and qubit-efficient solver for the LABS problem},
  author={Sciorilli, Marco and Camilo, Giancarlo and Maciel, Thiago O and Canabarro, Askery and Borges, Lucas and Aolita, Leandro},
  journal={arXiv preprint arXiv:2506.17391},
  year={2025}
}

@article{zhang25,
  title={New Improvements in Solving Large LABS Instances Using Massively Parallelizable Memetic Tabu Search},
  author={Zhang, Zhiwei and Shen, Jiayu and Kumar, Niraj and Pistoia, Marco},
  journal={arXiv preprint arXiv:2504.00987},
  year={2025}
}

@article{shaydulin24,
  title={Evidence of scaling advantage for the quantum approximate optimization algorithm on a classically intractable problem},
  author={Shaydulin, Ruslan and Li, Changhao and Chakrabarti, Shouvanik and DeCross, Matthew and Herman, Dylan and Kumar, Niraj and Larson, Jeffrey and Lykov, Danylo and Minssen, Pierre and Sun, Yue and others},
  journal={Science Advances},
  volume={10},
  number={22},
  pages={eadm6761},
  year={2024},
  publisher={American Association for the Advancement of Science}
}

@article{kettunen97,
  title={{Code selection for CDMA systems}},
  author={Kettunen, Kimmo},
  journal={Department of Information Studies, University of Tampere, Finland},
  year={1997}
}

@article{schmidt16,
author = {Schmidt, Kai-Uwe},
year = {2016},
month = {01},
pages = {},
title = {Sequences with small correlation},
volume = {78},
journal = {Designs, Codes and Cryptography},
doi = {10.1007/s10623-015-0154-7}
}

@incollection{mullen13,
    author = {Mullen, Gary L. and Panario, Daniel},
    title = {Other correlation measures},
    chapter = {10.3.5},
    booktitle = {Handbook of Finite Fields},
    isbn = {143987378X},
    publisher = {Chapman \& Hall/CRC},
    year =  {2013},
    edition = {1st},
    pages = {322-324}, 
}

@article{brest25,
  title={Oddaljenost neperiodičnih binarnih zaporedij glede na dve meri avtokorelacijskih lastnosti},
  author={Brest, Janez and Brest, Aljaž and Pšeničnik, Blaž and Popič, Jan and Bosković Borko},
  journal={Elektrotehniski Vestnik},
  volume={92},
  number={3},
  pages={97--103},
  year={2025},
  publisher={Elektrotehniski Vestnik},
  note={(In Slovene)}
}

@article{healy24,
  author       = {Healy, John and McInnes, Leland},
  title        = {Uniform manifold approximation and projection},
  journal = {Nature Reviews Methods Primers},
  year         = {2024},
  date         = {2024-11-21},
  volume       = {4},
  number       = {1},
  pages        = {82},
  issn         = {2662-8449},
  doi          = {10.1038/s43586-024-00363-x},
  url          = {https://doi.org/10.1038/s43586-024-00363-x}
}

@article{oliveira99,
doi = {10.1088/0305-4470/32/50/302},
year = {1999},
month = {12},
publisher = {},
volume = {32},
number = {50},
pages = {8793},
author = {Viviane M de Oliveira and José F Fontanari and Peter F Stadler},
title = {Metastable states in short-ranged p-spin glasses},
journal = {Journal of Physics A: Mathematical and General},
}

@article{packebusch16,
  title={Low autocorrelation binary sequences},
  author={Packebusch, Tom and Mertens, Stephan},
  journal={Journal of Physics A: Mathematical and Theoretical},
  volume={49},
  number={16},
  pages={165001},
  year={2016},
  publisher={IOP Publishing}
}

@article{mertens96,
  title={Exhaustive search for low-autocorrelation binary sequences},
  author={Mertens, Stephan},
  journal={Journal of Physics A: Mathematical and General},
  volume={29},
  number={18},
  pages={L473},
  year={1996},
  publisher={IOP Publishing}
}

@article{prestwich07,
  author       = {Steven Prestwich},
  title        = {Exploiting Relaxation in Local Search for LABS},
  journal      = {Annals of Operations Research},
  volume       = {156},
  pages        = {129--141},
  year         = {2007},
  month        = {12},
  doi          = {10.1007/s10479-007-0226-9},
  publisher    = {Springer},
  OPTurl          = {https://doi.org/10.1007/s10479-007-0226-9}
}

@article{prestwich13,
  title={Improved branch-and-bound for low autocorrelation binary sequences},
  author={Prestwich, Steven D},
  journal={arXiv preprint arXiv:1305.6187},
  year={2013}
}

@misc{wiggenbrock10,
  author       = {Jens Wiggenbrock},
  title        = {Parallele Optimierungsstrategien des LABS-Problems in einem GPU-Grid},
  type         = {Bachelor's thesis},
  school       = {Fachhochschule Südwestfalen},
  year         = {2010},
  language     = {German},
  note={(In German)}
}

@article{browein04,
  author={Borwein, P. and Choi, K.-K.S. and Jedwab, J.},
  journal={IEEE Transactions on Information Theory}, 
  title={Binary sequences with merit factor greater than 6.34}, 
  year={2004},
  volume={50},
  number={12},
  pages={3234-3249},
  doi={10.1109/TIT.2004.838341}}

@article{hoholdt88,
  author={Hoholdt, T. and Jensen, H.E.},
  journal={IEEE Transactions on Information Theory}, 
  title={Determination of the merit factor of Legendre sequences}, 
  year={1988},
  volume={34},
  number={1},
  pages={161-164},
  doi={10.1109/18.2620}}

@article{baden11,
    author = {Baden, John Michael},
    journal = {IEEE Transactions on Information Theory},
    title = {{Efficient Optimization of the Merit Factor of Long Binary Sequences}},
    year = {2011},
    volume = {57},
    number = {12},
    pages = {8084-8094},
    doi = {10.1109/TIT.2011.2164778},
}

@article{jedwab13,
    title = {Advances in the merit factor problem for binary sequences},
    journal = {Journal of Combinatorial Theory, Series A},
    volume = {120},
    number = {4},
    pages = {882-906},
    year = {2013},
    doi = {10.1016/j.jcta.2013.01.010},
    author = {Jonathan Jedwab and Daniel J. Katz and Kai-Uwe Schmidt},
}

@inproceedings{farnane18,
    author = {Farnane, Kaoutar and Minaoui, Khalid and Aboutajdine, Driss},
    booktitle = {2018 4th International Conference on Optimization and Applications (ICOA)},
    title = {Local search algorithm for low autocorrelation binary sequences},
    year = {2018},
    volume = {},
    number = {},
    pages = {1-5},
    doi = {10.1109/ICOA.2018.8370526}
}

@inproceedings{halim08,
    author = {Halim, Steven and Yap, Roland H. C. and Halim, Felix},
    editor = {Stuckey, Peter J.},
    title = {{Engineering Stochastic Local Search for the Low Autocorrelation Binary Sequence Problem}},
    booktitle = {Principles and Practice of Constraint Programming},
    year = {2008},
    publisher = {Springer Berlin Heidelberg},
    address = {Berlin, Heidelberg},
    pages = {640--645},
}

@article{gallardo09,
    title = {Finding low autocorrelation binary sequences with memetic algorithms},
    journal = {Applied Soft Computing},
    volume = {9},
    number = {4},
    pages = {1252-1262},
    year = {2009},
    doi = {10.1016/j.asoc.2009.03.005},
    author = {José E. Gallardo and Carlos Cotta and Antonio J. Fernández},
}

@article{boskovic17,
    title = {{Low-autocorrelation binary sequences: On improved merit factors and runtime predictions to achieve them}},
    journal = {Applied Soft Computing},
    volume = {56},
    pages = {262-285},
    year = {2017},
    issn = {1568-4946},
    doi = {10.1016/j.asoc.2017.02.024},
    author = {Borko Bošković and Franc Brglez and Janez Brest},
}

@article{brest18,
    author = {Brest, Janez and Bošković, Borko},
    journal = {IEEE Access},
    title = {{A Heuristic Algorithm for a Low Autocorrelation Binary Sequence Problem With Odd Length and High Merit
Factor}},
    year = {2018},
    volume = {6},
    number = {},
    pages = {4127-4134},
    doi = {10.1109/ACCESS.2018.2789916}
}

@article{brest22,
    title={{Computational Search of Long Skew-symmetric Binary Sequences with High Merit Factors}},
    volume={28},
    DOI={10.13164/mendel.2022.2.017},
    number={2},
    journal={MENDEL},
    author={Brest, Janez and Bošković, Borko},
    year={2022},
    month={12},
    pages={17-24}
}

@article{herzog23,
title = {Analysis based on statistical distributions: A practical approach for stochastic solvers using discrete and continuous problems},
journal = {Information Sciences},
volume = {633},
pages = {469-490},
year = {2023},
issn = {0020-0255},
doi = {https://doi.org/10.1016/j.ins.2023.03.081},
OPTurl = {https://www.sciencedirect.com/science/article/pii/S0020025523003869},
author = {Jana Herzog and Janez Brest and Borko Bošković},
}

@article{zurek17,
title = {Toward hybrid platform for evolutionary computations of hard discrete problems},
journal = {Procedia Computer Science},
volume = {108},
pages = {877-886},
year = {2017},
issn = {1877-0509},
doi = {doi.org/10.1016/j.procs.2017.05.201},
author = {Dominik Żurek and Kamil Piętak and Marcin Pietroń and Marek Kisiel-Dorohinicki}
}

@article{pietak19,
title = {{Striving for performance of discrete optimisation via memetic agent-based systems in a hybrid CPU/GPU environment}},
journal = {Journal of Computational Science},
volume = {31},
pages = {151-162},
year = {2019},
issn = {1877-7503},
doi = {doi.org/10.1016/j.jocs.2019.01.007},
author = {Kamil Piętak and Dominik Żurek and Marcin Pietroń and Andrzej Dymara and Marek Kisiel-Dorohinicki},
}

@inproceedings{zurek22,
  title={A deep neural network as a TABU support in solving LABS problem},
  author={{\.Z}urek, Dominik and Pietro{\'n}, Marcin and Pi{\k{e}}tak, Kamil and Kisiel-Dorohinicki, Marek},
  booktitle={International Conference on Computational Science},
  pages={237--243},
  year={2022},
  organization={Springer}
}

@article{cadavid25,
  title={Scaling advantage with quantum-enhanced memetic tabu search for LABS},
  author={Cadavid, Alejandro Gomez and Chandarana, Pranav and Romero, Sebasti{\'a}n V and Trautmann, Jan and Solano, Enrique and Patti, Taylor Lee and Hegade, Narendra N},
  journal={arXiv preprint arXiv:2511.04553},
  year={2025},
  doi={https://doi.org/10.48550/arXiv.2511.04553}
}

@misc{eppstein15,
  author  = {Eppstein, David},
  title   = {PADS: Python Algorithms and Data Structures},
  url     = {http://www.ics.uci.edu/~eppstein/PADS/},
  year    = {2015},
  note    = {MIT licensed software library}
}

@article{skula26,
  author  = {Skula, Milan and Pies, Martin and Hajovsky, Radovan and Velicka, Jan and Vala, David},
  title   = {Multi-criteria selection of a synchronisation word for low-power IoT receivers based on the IQRF standard},
  journal = {Scientific Reports},
  year    = {2026},
  date    = {2026-02-13},
  volume  = {16},
  number  = {1},
  pages   = {8777},
  doi     = {10.1038/s41598-026-38142-1},
  url     = {https://doi.org/10.1038/s41598-026-38142-1}
}
